\documentclass[sigconf,nonacm]{modified}

\AtBeginDocument{%
  }

\setcopyright{none}
\begin{document}

\title{Stance Detection and Open Research Avenues}


\author{Dilek K\"u\c{c}\"uk}
%\authornote{The authors will not be able to attend the conference in person.}
\email{dilek.kucuk@tubitak.gov.tr}
\affiliation{%
  \institution{T\"{U}B\.ITAK Marmara Research Center}
  \streetaddress{METU Campus}
  \city{Ankara}
  \country{Turkey}}

\author{Fazli Can}
\email{canf@cs.bilkent.edu.tr}
\affiliation{%
  \institution{Bilkent University}
  \streetaddress{Bilkent Campus}
  \city{Ankara}
  \country{Turkey}
}
\renewcommand{\shortauthors}{K\"u\c{c}\"uk and Can}

\begin{abstract}
This tutorial aims to cover the state-of-the-art on stance detection and address open research avenues for interested researchers and practitioners. Stance detection is a recent research topic where the stance towards a given target or target set is determined based on the given content and there are significant application opportunities of stance detection in various domains. The tutorial comprises two parts where the first part outlines the fundamental concepts, problems, approaches, and resources of stance detection, while the second part covers open research avenues and application areas of stance detection. The tutorial will be a useful guide for researchers and practitioners of stance detection, social media analysis, information retrieval, and natural language processing.
\end{abstract}



\maketitle

\section{Introduction}

Stance detection is a research problem focusing on people's positions towards specific targets, in natural language texts \cite{kuccuk2020stance,kuccuk2021stance,kuccuk2022tutorial}. It can be considered as a subproblem of affective computing, along with related research topics such as sentiment analysis. Stance classification, stance analysis, and stance extraction are also used to refer to the problem of stance detection in the related literature.

One of the important milestones of stance detection research is the stance detection shared task on English tweets performed in 2016 \cite{mohammad2016semeval}. Within the course of this competition, a stance-annotated tweet dataset is created and publicly shared \cite{mohammad2016dataset}. This shared task is followed by other similar competitions on texts in Chinese \cite{xu2016chinesestancetask}, Spanish and Catalan \cite{taule2017stance}, Italian \cite{cignarella2020sardistance}, and Basque \cite{agerri2021vaxxstance}.

\begin{figure}[h]
  \centering
  \includegraphics[width=\linewidth]{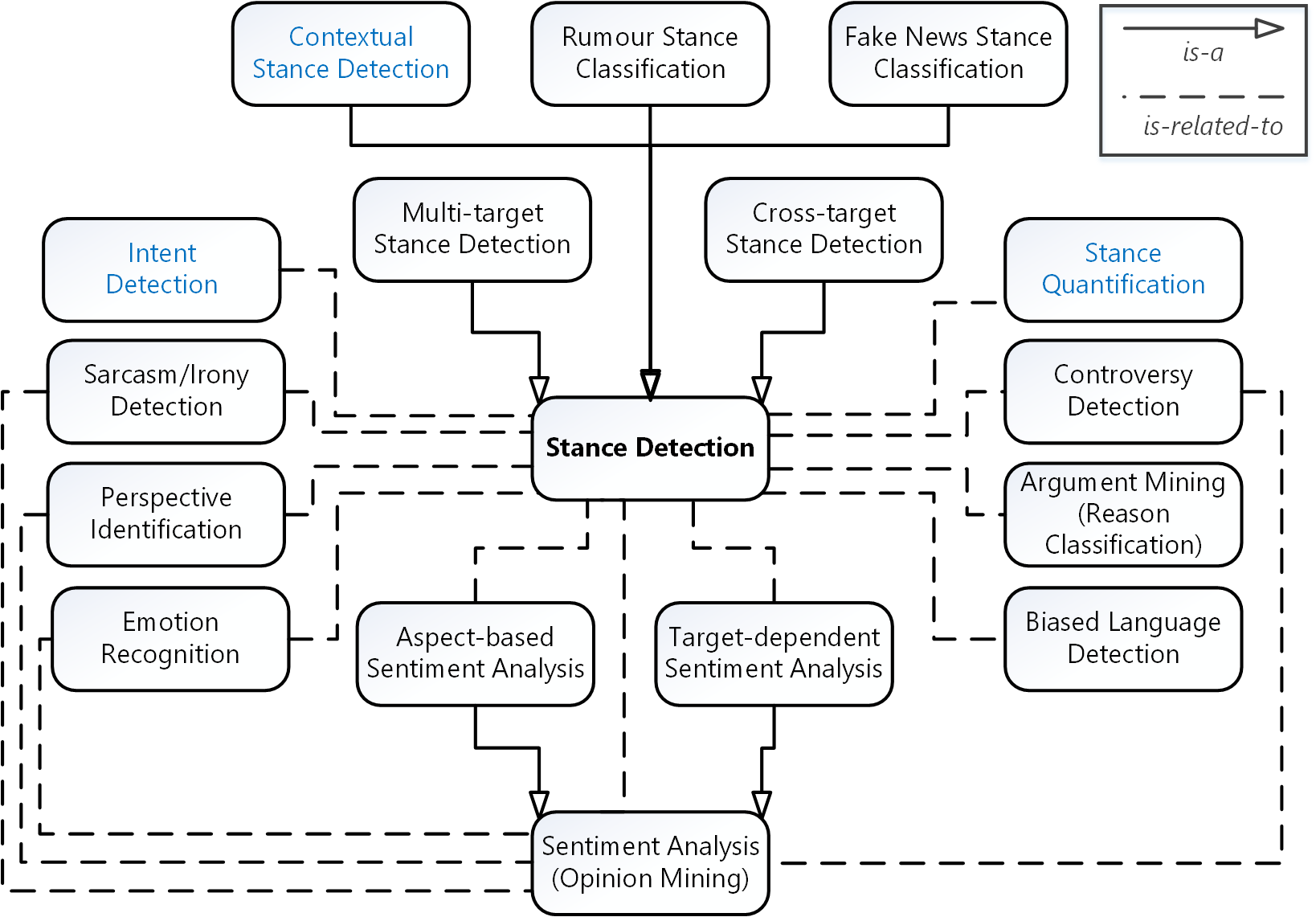}
  \caption{Stance Detection, Its Subproblems, and Related Research Problems (Revised Version of the Corresponding Figure in \cite{kuccuk2020stance}).}
  \label{fig1}
  \Description{Revised Schematic Representation of Stance Detection, Its Subproblems, and Related Research Problems.}
\end{figure}

Significant subproblems of stance detection and closely-related problems have been previously discussed in details in \cite{kuccuk2020stance,kuccuk2021stance,kuccuk2022tutorial}. Based on the related figure in \cite{kuccuk2020stance}, a revised schematic representation demonstrating stance detection, its subproblems, and related problems is given in Figure \ref{fig1}. The newly-added problems are \textit{contextual stance detection} \cite{cignarella2020sardistance,aldayel2021stance}, \emph{intent detection} \cite{kuccuk2021sentiment}, and \textit{stance quantification} \cite{kuccuk2021stancequantification}, which are shown in blue in the figure.

\textit{Contextual stance detection} does not only use the input text, but makes use of contextual information (social media user profiles, interactions between posts, etc.) as well during the stance detection procedure. Hence, contextual stance detection is a subproblem of generic stance detection. \emph{Intent detection} (together with slot filling), on the other hand, is a research problem of dialogue systems and natural language understanding, where the goals of the users are extracted from their utterances \cite{niu2019novel}. \textit{Stance quantification} aims the determine the percentages of the textual items belonging to distinct stance classes, instead of labeling each item with its stance label \cite{kuccuk2021stancequantification}. We aim to cover all of the research problems in Figure \ref{fig1} during our tutorial.

Most of the recent work on stance detection employs deep learning approaches \cite{zhao2020pretrained,schiller2021stance}. Yet, ensemble methods are also commonly observed in stance detection research \cite{chen2021integrating}.

It has been previously reported that stance detection studies were published on several languages including English, Chinese, Spanish, Catalan, Italian, Japanese, Turkish, Czech, Russian, and Arabic \cite{kuccuk2020stance}. Recent work also reveal that there are related studies performed on Basque \cite{agerri2021vaxxstance}, German \cite{mascarell2021stance}, Portuguese \cite{won2022ss}, and Persian \cite{nasiri2022persian}, too.

This stance detection tutorial will consist of two main parts where the first part will be devoted to presentation of basic concepts, related research problems, stance detection competitions, machine learning and deep learning based approaches, and stance detection resources like the datasets, with particular emphasis on publicly shared datasets. The second part of the tutorial will mostly cover open research topics related to stance detection. Basically the following open research topics will be considered in the second part:

\begin{enumerate}

	\item	\emph{Stance Detection in Data Streams}: Data streams constitute a significant research area \cite{bonab2016theoretical,gozuaccik2021concept,gulcan2022unsupervised} and stance detection can also be applied to large volumes of particularly social media posts provided as streams. Hence, stance detection in data streams is one of the fruitful open research areas of stance detection \cite{bechini2021addressing}.

	\item \emph{Finer Grained Stance Detection}: Most of the work on stance detection performs three-way or two-way classification using the stance classes of \emph{favor}, \emph{against}, \emph{none}, and \emph{neutral}. However, considering related work on sentiment analysis, corresponding polarity classes occasionally include classes that posses additional semantic information such as strongly positive, strongly negative, weakly positive etc. In order to extract richer stance-related information from the underlying texts, finer-grained stance detection can be employed similarly: by extending the basic stance label set with new labels such as strongly favor, weakly favor, strongly against, etc.

	\item \emph{Stance Detection on Legal Texts and Other Text Genres}: Recently, most of the stance detection research has been conducted on social media texts, particularly on tweets. But stance detection can also be performed on legal texts and on other text genres such as generic news articles \cite{mascarell2021stance} which may lead to fruitful results for the corresponding domains.
	
	\item \emph{Cross-lingual and Multilingual Stance Detection}: In cross-lingual stance detection, stance-annotated dataset in one language is used to improve stance detection in another language \cite{mohtarami2019contrastive,hardalov2022crosslingual} where the latter language is usually a low-resource one. On the other hand, in multilingual stance detection \cite{lai2020multilingual}, a stance detection approach is usually applied to datasets in different languages to test the portability of the employed approach. Both cross-lingual and multilingual stance detection are part of important open research topics of stance detection.

\end{enumerate}

Second part of the tutorial will also cover important application areas of stance detection such as topics in information retrieval \cite{al2021novel}, fake news detection \cite{shu2019beyond}, and rumour detection and resolution \cite{rumour2022,huang2022stance}.

\section{Target Audience, Prerequisites, and Benefits}

The main two benefits of this tutorial are listed below:

\begin{itemize}
    \item	The tutorial will firstly cover the basic concepts, approaches, and resources for stance detection. While covering approaches and resources such as datasets, more emphasis will be given to recent research on stance detection.
    \item	Secondly, significant open research topics related to stance detection will be outlined, along with the main application areas of stance detection. Although these two aspects can be thought as intersecting, here we will put more emphasis on open research avenues of stance detection.
\end{itemize}

Target audience of this tutorial include researchers and practitioners interested in stance detection, social media analysis, affective computing, natural language processing, and information retrieval. We believe that the tutorial will be beneficial to those researchers\slash practitioners who have prior information about stance detection as well as to those who do not have, since the tutorial will cover basics, current state-of-the-art, and significant future research opportunities.

There are no prerequisites for this tutorial.

\section{Tutorial Outline}

Below provided is the outline of our half-day tutorial proposal on stance detection and open research problems related to stance detection.

\begin{enumerate}
	\item	Tutorial Part I: Basics, Competitions, Approaches, and Datasets
		\begin{enumerate}	
			\item	Basic concepts of stance detection
				\begin{enumerate}	
					\item	Stance detection: problem definition
					\item	Subproblems of stance detection
					\item	Problems related to stance detection
				\end{enumerate}
			\item	Stance detection competitions (shared tasks)
				\begin{enumerate}	
					\item	Shared task on English tweets (2016)
					\item	Shared task on Chinese microblogs (2016)
					\item	Shared task on Spanish and Catalan tweets (2017)
					\item	Shared task on Italian tweets (2020)
					\item	Shared task on Basque tweets (2021)
				\end{enumerate}
			\item	Approaches to stance detection
			\item	Stance detection datasets
		\end{enumerate}
	\item Tutorial Part II: Open Research Avenues and Application Areas
		\begin{enumerate}
			\item	Open research avenues		
				\begin{enumerate}
					\item	Stance detection in data streams		
					\item	Finer grained stance detection		
					\item	Stance detection on legal documents and on other text genres		
					\item	Cross-lingual and multilingual stance detection	
				\end{enumerate}
			\item	Application areas		
				\begin{enumerate}
					\item	Information retrieval		
					\item	Rumour classification		
					\item	Fake news detection
				\end{enumerate}
		\end{enumerate}
\end{enumerate}

\section{Previous Related Tutorials}

\subsection{Detection and Characterization of Stance on Social Media (ICWSM-2020)}

This stance detection tutorial was performed as part of the ICWSM-2020 conference and particularly considers stance detection on social media\footnote{\url{http://smash.inf.ed.ac.uk/files/Part2_phase2.pdf}}. Our current stance detection tutorial covers more recent work, is not limited to stance detection on social media only (though we should acknowledge that most of the related work is performed on social media), and finally allots almost half of the tutorial duration to open research avenues. Therefore, we believe that our tutorial will be very beneficial for researchers and graduate students who are about to start stance detection research.

\subsection{Stance Detection: Concepts, Approaches, Resources, and Outstanding Issues (SIGIR-2021)}

This tutorial was presented at SIGIR-2021 conference and constitutes the initial form of our tutorials on stance detection \cite{kuccuk2021stance}. The tutorial is mostly built upon the work presented in \cite{kuccuk2020stance} and the current tutorial has intersecting content with this preceding tutorial, particularly regarding the basic concepts of stance detection. Yet, as emphasized in the previous sections, the current tutorial covers research problems very much related to stance detection that are not considered in this preceding tutorial. Again, being a more recent tutorial, our current tutorial includes more recent studies on stance detection and emphasizes open research topics.

\subsection{A Tutorial on Stance Detection (WSDM-2022)}

This is the second version of our tutorials on stance detection and was presented at WSDM-2022 \cite{kuccuk2022tutorial}. It is an extended form of the initial tutorial presented at SIGIR-2021, including more recent related studies compared to the previous one. Hence, the current tutorial differs from this second form in that the current one covers more recent research and additionally emphasizes important open research avenues.

\section{Short Biographies of Presenters}

\subsection{Dilek K\"u\c{c}\"uk}

Dilek K\"u\c{c}\"uk is an associate professor and senior chief researcher at T\"{U}B\.ITAK Marmara Research Center (MRC) in Ankara, Turkey. She received her B.Sc., M.Sc. and Ph.D. degrees in Computer Engineering all from Middle East Technical University (Ankara, Turkey) in 2003, 2005, and 2011, respectively. From May 2013 to May 2014, she studied as a post-doctoral researcher at European Commission's Joint Research Centre in Italy. Her research interests include stance detection, social media analysis, and energy informatics. She is the author or co-author of 16 papers published in SCI-indexed journals (including ACM-CSUR and IEEE transactions) and 42 papers presented at international conferences/workshops. She was a joint tutorial presenter at SIGIR-2021 and WSDM-2022 conferences. Her personal Web page is available at \url{https://dkucuk.github.io/en.html}.

\subsection{Fazli Can}

Fazli Can received the B.Sc. and M.Sc. degrees in Electrical and Electronics and Computer Engineering and the Ph.D. degree in Computer Engineering from Middle East Technical University, Ankara, Turkey, in 1976, 1979, and 1985, respectively. He conducted his Ph.D. research under the supervision of Prof. E. Ozkarahan; at Arizona State University, Tempe, AZ, USA, and Intel, Chandler, AZ, USA; as a part of the RAP Database Machine Project. He is currently a faculty member at Bilkent University, Ankara. Before joining Bilkent, he was a tenured full professor at Miami University, Oxford, OH, USA. He co-edited ACM SIGIR Forum from 1995 to 2002 and is a co-founder of the Bilkent Information Retrieval Group, Bilkent University. His interest in dynamic information processing dates back to his 1993 incremental clustering paper in ACM Transactions on Information Systems and some other earlier work with Prof. E. Ozkarahan on dynamic cluster maintenance. His current research interests include information retrieval and data mining. His personal Web page is available at \url{http://www.cs.bilkent.edu.tr/~canf}.


\end{document}